# A Multi-Agent Approach for Adaptive Finger Cooperation in Learning-based In-Hand Manipulation

Lingfeng Tao\*, Jiucai Zhang^, Michael Bowman \*, and Xiaoli Zhang\*, *Member, IEEE*

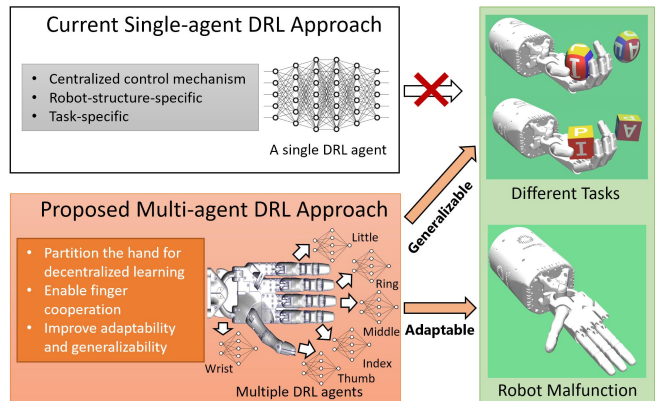

Figure 1. The finger-based multi-agent DRL approach partitions the hand structure and assign each agent to control a specific portion of the hand (5 fingers and a wrist in our case). The agents can learn to cooperate to complete the task. This multi-agent approach is generalizable to different tasks and the trained policies have higher adaptability to robot malfunction.

*Abstract—* In-hand manipulation is challenging for a multi-finger robotic hand due to its high degrees of freedom and the complex interaction with the object. To enable in-hand manipulation, existing deep reinforcement learning based approaches mainly focus on training a single robot-structure-specific policy through the centralized learning mechanism, lacking adaptability to changes like robot malfunction. To solve this limitation, this work treats each finger as an individual agent and trains multiple agents to control their assigned fingers to complete the in-hand manipulation task cooperatively. We propose the Multi-Agent Global-Observation Critic and Local-Observation Actor (MAGCLA) method, where the critic can observe all agents' actions globally, and the actor only locally observes its neighbors' actions. Besides, conventional individual experience replay may cause unstable cooperation due to the asynchronous performance increment of each agent, which is critical for in-hand manipulation tasks. To solve this issue, we propose the Synchronized Hindsight Experience Replay (SHER) method to synchronize and efficiently reuse the replayed experience across all agents. The methods are evaluated in two in-hand manipulation tasks on the Shadow dexterous hand. The results show that SHER helps MAGCLA achieve comparable learning efficiency to a single policy, and the MAGCLA approach is more generalizable in different tasks. The trained policies have higher adaptability in the robot malfunction test compared to the baseline multi-agent and single-agent approaches.

## I. INTRODUCTION

Dexterous robot hands have the high potential to enable in-hand manipulation, which is one of the essential functions for manufacturing [1], assembly [2], and assistive living [3]. The rapid development of miniaturized actuators and sensors has provided hardware foundations for dexterous robotic hands, such as the Shadow hand [4], an anthropomorphic robotic hand with 24 degrees of freedom (DoFs), in which 20 joints are independently controllable. It has been used in complex in-hand manipulation tasks such as solving a Rubik's Cube [5]. With the readiness of robot hardware, researchers have been putting efforts into developing generalizable and adaptable methods for in-hand manipulation applications.

Deep reinforcement learning (DRL) [6] has shown its abilities in recent research [4, 5, 7] to solve dexterous in-hand manipulation tasks thanks to its learning capability, which enables the robot to find an autonomous control policy by interacting with the task environment through exploration and exploitation. However, current DRL-based approaches for in-hand manipulation only train a single policy with the global observation of the whole environment as the input and outputs global actions to control the entire robot hand to interact with the object. As a result, the policy becomes robot-structure-specific and object-specific, lacking adaptability to changes [8], such as robot malfunction. Because the changes will affect the whole policy, leading to performance reduction or task failure. Current efforts [4, 5] to improve the adaptability of the single DRL policy for in-hand manipulation rely on adding randomized noise to the sensor, control signal, and appearance if using visual input. These approaches help the DRL policy adapt to the uncertainty and disturbance in the environment but do not adapt to changes like robot malfunction.

Multi-Agent DRL has shown high adaptability in multiple robot control tasks [9-11] because of the decentralized learning approach, which improves the system flexibility and resilience [12-14] by limiting the influence of the changes on local agents instead of the whole system. Literature [15][16] has shown that a multi-agent DRL setup can control multiple robot arms individually by complete manipulation tasks like picking up objects. Similarly, we can treat each finger of the robot hand as a small robot arm because each finger can be individually controlled. As an inspiration, this work proposes to consider robot fingers as individual manipulators that cooperate to complete the manipulation task (Fig. 1). To the authors' knowledge, it is the first time solving in-hand manipulation with the multi-agent DRL approach.

We propose the Multi-Agent Global-observation Critic and Local-observation Actor (MAGCLA), which belongs to the category of Centralized Training and Decentralized Execution (CTDE) [17] method. CTDE is widely adopted in multi-agent actor-critic DRL methods. The rationale of CTDE is that the critic can observe extra information to benefit the actor's

This material is based on work supported by the US NSF under grant 1652454 and 2114464. Any opinions, findings, conclusions, or recommendations expressed in this material are those of the authors and do not necessarily reflect those of the National Science Foundation.

\*L. Tao, M. Bowman, and X. Zhang are with Colorado School of Mines, Intelligent Robotics and Systems Lab, 1500 Illinois St, Golden, CO 80401 (e-mail: tao@mines.edu, mibowman@mines.edu, xlzhang@mines.edu).

^J. Zhang is with the GAC R&D Center Silicon Valley, Sunnyvale, CA 94085 USA (e-mail: zhangjiucai@gmail.com).

training while the actor does not need the extra information in execution. In MAGCLA, the Global-observation critic means that in addition to the environment state, the critic of an agent also has a global observation of all other agents' actions. Compared to conventional CTDE methods, the uniqueness of MAGCLA is that the actor can observe its neighbor's actions instead of its own state only. Observing neighbors' information helps agents learn cooperative behavior in a multi-agent setup [18-20], which is critical for in-hand manipulation.

As with most DRL approaches, MAGCLA relies on a replay buffer to manage the exploration experience. In the multi-agent in-hand manipulation setup, if updating each agent with individually sampled experience as in conventional multi-agent approaches, the agents' performance increments may not follow the same pace, causing unstable training and converging to low-performance policies. We propose the Synchronized Hindsight Experience Relay (SHER) to solve this issue, expanding the HER method [21] to a multi-agent setup. SHER synchronizes the experience sampling across all agents to ensure that the agents' performance increments are at the same pace. SHER works well for all off-policy multi-agent DRL approaches in in-hand manipulation tasks. In summary, the contributions of this work are:

1) Model the in-hand manipulation task as a finger-based multi-agent setup for the first time.
2) Develop a MAGCLA method to enable finger cooperation in in-hand manipulation.
3) Develop a SHER method for stable performance increments by synchronizing the experience sampling across all agents.
4) Validate the MAGCLA and SHER methods on the Shadow dexterous hand in two in-hand manipulation tasks and compare their generalizability and adaptability with the existing single-agent and multi-agent DRL approaches.

## II. RELATED WORK

### A. Learning-based In-Hand Manipulation

Conventional analytical control methods rely on solving partial-differential kinematics equations [22] or optimizing toward objective functions [23], bringing high computational costs with complex robot hand structures. Thus, real-time processing [24] is usually challenging with analytical methods. The complex interaction with the object also makes the manipulation task difficult for the analytical methods.

Single-agent DRL methods have demonstrated their capability to handle in-hand manipulation tasks [4, 5]. The OpenAI Gym [25] toolkit implements challenging in-hand manipulation tasks [26] with the Shadow robot hand as a standard benchmark. With the maturity of single-agent DRL-based in-hand manipulation, recent literature focuses on implementing the DRL agent trained in simulation to the physical robot hand to complete real-world tasks, such as solving a Rubik's cube [5] or rotating a block to a target pose [4]. Randomization [27] is applied to sensing, actuation, and appearance to improve the policy adaptability to noise and disturbance. Learn from demonstration methods [28] are also used to improve training efficiency by initializing the DRL policy for in-hand manipulation tasks such as rotating door handles and using a hammer or screwdriver.

### B. Multi-Agent Approach in Learning-based Robot Training

Multi-agent DRL methods have been adopted in complex multiple robot control [9][10] for their high adaptability, thanks to their decentralized learning mechanism. In [16], a dual-arm multi-agent DRL approach was proposed to solve cooperative grasping tasks. Another dual-arm setup is reported in [29] to solve the table-carrying task. Recently, a multi-arm DRL motion planner was proposed to generate a trajectory for an 8-arm system to reach its target end-effector poses [15]. Although, the principle of in-hand manipulation is like multi-arm cooperation, where the fingers cooperate to complete the manipulation. However, the multi-agent DRL methods are not yet studied for in-hand manipulation.

### C. Centralized Training and Decentralized Execution

CTDE is originally developed for simulation to real-world transfer applications [30], allowing high dimension input for the critic network and low dimensional observation input for the actor, so the critic can benefit from more information while training in simulation, and the actor can adapt to the sparse information in the real world. CTDE was widely adopted in multi-agent actor-critic DRL approaches to improve learning stability and policy performance. CTDE allows the agents' critics to observe extra information of other agents during the training and only use the actors as control policies in testing without such extra information. MADDPG [31] and COMA [32] are well-known CTDE methods in multi-agent actor-critic. The proposed MAGCLA is more like MADDPG because each agent in MAGCLA has its own actor and critic rather than a shared critic in COMA. The difference between the proposed MAGCLA and MADDPG is that the actor of an agent in MAGCLA can still observe the actions of its neighbors instead of only its own state. The extra neighbor information helps the agents to understand their interactions and learn cooperative behaviors.

### D. Advanced Experience Replay Strategy

Advanced experience-replay strategies like HER, Prioritized Experience Replay [33], and their derivatives have been proposed to reuse the experience efficiently. HER is a post-experience editing method proposed to accelerate single-agent training in target-based tasks, such as pushing, sliding, pick-and-place [21], and in-hand manipulation [26]. The rationale of HER is that when the agent is exploring a goal $G$ and collecting a trajectory $(s_1, ..., s_T)$, where $s$ is the state, it may end up with a state $s_T$ that is not the target, making the trajectory a failed exploration. However, we can edit the trajectory in the replay buffer by treating the last state $s_T$ as a new goal $G'$ and resample the reward based on the transition to the new goal $G'$. Then the failed trajectory becomes a successful trajectory which can be used for policy updating. In this work, the proposed SHER method adopts the HER strategy and expands it to the multi-agent setup with an experience synchronization approach to help the agents to learn cooperative behaviors.

## III. METHODOLOGY

This section introduces the modeling of the multi-agent in-hand manipulation and the agent representation in III. A. The development of the MAGCLA is explained in III. B. The SHER method is shown in III. C.

### A. Multi-agent Modeling and Representation

We model the multi-agent in-hand manipulation as a Markov game [34], a multi-agent extension of the Markov Decision Process [35]. The Markov game contains $N$ agents, a set of action space $A_1, ..., A_N$, and a set of observations $O_1, ..., O_N$

**Algorithm 1** MAGCLA and SHER

**Initialize** critic $Q_i^\mu$, actor $\mu_{\theta_i}$, replay buffer $D$, and random noise $\mathcal{N}$.
1: **for** episode = 1 to $M$ **do**
2:     Initialize state $x$
3:     **for** $t = 1$ to max-episode-length **do**
4:         for agent $i$, selection action $a_i = \mu_{\theta_i}(o_{actor_i}) + \mathcal{N}$
5:         Collect transition $(x, a_1, \ldots, a_N, r, x')$ to replay buffer $D$
6:         Set $x \leftarrow x'$
7:         **for** agent $i = 1$ to $N$ **do**
8:             Select shared start state $x_\alpha$
9:             Synchronize HER minibatches for all agents with $x_\alpha$
10:           Update critic $Q_i^\mu$ and actor $\mu_{\theta_i}$ with eq. 4 and eq. 5
11:         **end for**
12:         Update target network parameters $\theta_i'$ with eq. 6
13:     **end for**
14: **end for**

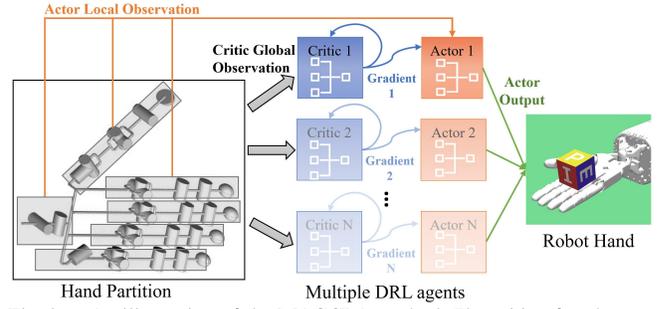

Fig. 2. An illustration of the MAGCLA method. The critic of each agent has a global observation to the robot hand and all actions of other agents. The actor of each agent can observe the actions of its neighbors. For simplicity, in the figure, we only draw the actor local observation for the thumb agent.

that are assigned to each agent. Each agent follows a policy $\pi_{\theta_i}: O_i \times A_i \mapsto [0,1]$. A state $S$ is defined to describe the Markov game. The execution of all agents' actions produces the transition to the next state by following the state transition function $\Gamma: S \times A_1 \times \ldots \times A_N \mapsto S'$. A reward function is designed for each agent based on the state and action $r_i: \times A_i \mapsto \mathbb{R}$. The agent $i$ should maximize its expected total reward $R_i = \sum_{t=0}^T \gamma^t r_i^t$, where $\gamma$ is a discount factor, $t$ is the time step, and $T$ is the maximum time steps.

For multi-agent in-hand manipulation, the action space $A_i$ of agent $i$ is based on the hand partitions (Fig. 2). In this work, we assign each agent to control one of the fingers and an additional agent to control the wrist. The rationale for not assigning more agents to control each motor or fewer agents to control two or more fingers is that the aim is to maximize the independence of each agent and keep its functionality as a manipulator. The wrist agent controls the palm to adjust the hand pose during the manipulation process. The observable state helps the agent assess the state information in the manipulation process. The observable state is denoted as $x$, including the positions and velocities of the robot's joints and the Cartesian position and rotation of the object represented by a quaternion as its linear and angular velocities. The action and state spaces are normalized to -1 to 1 for stable training and to avoid overfitting.

*B. Global-Observation Critic and Local-Observation Actor*

MAGCLA uses a centralized critic for agent $i$ to approximate the action-value function $Q_i$ with a global observation
$$o_{critic} = (x, a_1, \ldots, a_N) \quad (1)$$
where $(a_1, \ldots, a_N)$ are the actions of all agents. The extra observation of all agents' actions in the critic can help the agent to understand their contribution to the manipulation task and how their action interacts with others and the object, enabling finger cooperation. For actor $i$, the observation is:
$$o_{actor_i} = (x, a_{i-1}, a_i, a_{i+1}) \quad (2)$$
which means that the actor can observe its neighbors' actions. It should be noted that $o_{actor_i}$ is flexible. In practice, the wrist actor observes all agents' actions. The thumb actor only observes the index, and the little actor only observes the ring.

MAGCLA adopts the deterministic policy gradient [36] method for continuous action space. For each agent $i$, we train a continuous actor $\mu_{\theta_i}$, where $\theta_i$ is the network parameters to maximize the objective function
$$J(\mu_{\theta_i}) = \mathbb{E}_{s \sim p^\pi}[R_i] \quad (3)$$
where $p^\pi$ is the state distribution. The gradient of the actor can be calculated as:

$$\nabla_{\theta_i} J(\mu_{\theta_i}) =$$
$$\mathbb{E}_{x, a \sim D}\left[\nabla_{\theta_i}\mu_{\theta_i}(a_i|o_{actor_i})\nabla_{a_i}Q_i^\mu(o_{critic})|_{a_i=\mu_{\theta_i}(o_{actor_i})}\right] \quad (4)$$

The gradient utilizes the extra information of all agents' actions in the critic observation to help the actor's training. $D$ is the replay buffer which contains a transition tuple $(x, x', a_1, \ldots, a_6, r_1, \ldots, r_6)$. The critic is updated by minimizing the loss function
$$\mathcal{L}(\theta_i) = \mathbb{E}_{x,a,r,x'}\left[Q_i^\mu(o_{critic}) - y\right]^2$$
$$\text{where } y = r_i + \gamma Q_i^{\mu'}(o'_{critic})|_{a'_i=\mu_{\theta'_i}(o'_{actor_i})} \quad (5)$$

$\mu'$ is the target policy with delayed parameters $\theta'$ for stable updating. Each agent's target policy $\theta_i'$ are updated at the end of every epoch as:
$$\theta_i' \leftarrow \tau\theta_i + (1-\tau)\theta_i' \quad (6)$$
where $\tau$ is the learning rate.

*C. Synchronized Hindsight Experience Replay*

HER has proved effective in helping train a single DRL agent for in-hand manipulation tasks. When directly implementing HER to multi-agent in-hand manipulation setup, each agent individually samples the experience from the replay buffer. After editing with HER, the agents will be updated with different trajectories with different goal positions (Fig. 3a). As a result, the policies may learn conflicting behaviors or have asynchronous performance improvements. This is acceptable for most multi-agent tasks as their policies are independent, where different performed agents cause less unstable factors to the training. The low-performed agent can quickly catch up as the training continues [37]. However, in-hand manipulation tasks require seamless cooperation between all agents. The task can easily fail because of mistakes made by weaker performing agents.

To solve the above issue, SHER synchronizes the replayed experience across all agents by selecting a shared start state

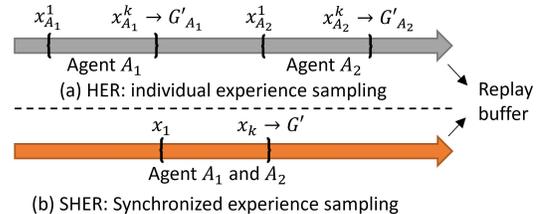

Fig. 3. The comparison between (a) HER and (b) SHER in a two-agent setup. $k$ is the sample length. In HER, each agent individually samples the experience, resulting in different goal state, causing asynchronous performance increments. In SHER, the sampled experience is synchronized for all agents. The agents update the policies toward the same goal, helping the agents learn cooperative behavior.

$x_\alpha$ for all agents, then samples the same period of experience with HER for all agents to update the policies (Fig. 3b). Since the agents share the same state transitions of the object, the SHER-edited trajectories will have the same new goal state. The synchronized experience helps to normalize the gradient across all the critics and keeps the performance increments of all agents at the same pace. The overall proposed MAGCLA and SHER are summarized in algorithm 1.

## IV. EXPERIMENTS

### A. Task Design

The MAGCLA-SHER approach requires a capable test platform to derive viable in-hand manipulation applications. The approach will be evaluated in a simulated environment for ease of training and testing. Specifically, we adopt the Shadow hand environments from the OpenAI GYM Robotics platform, which runs on the MuJoCo [38] physics simulator. We partition the Shadow hand into 6 agents, 5 agents control each finger, and the additional agent controls the wrist. Based on the hand partition, the 6 DRL agents are wrist (2 DoFs), thumb (5 DoFs), index (3 DoFs), middle (3 DoFs), ring (3 DoFs), and little (4 DoFs). Two in-hand manipulation environments (Fig. 4) are designed to evaluate the generalizability of our methods in different tasks:

1) *Block manipulation*. A block is placed on the Shadow hand's palm with a random initial pose. The task is to manipulate the block around the Z axis to achieve the target pose.
2) *Egg manipulation*. The task is similar to block manipulation, but an egg-shaped object is used.

In both tasks, a goal is achieved if the difference in the rotation is less than 0.1 rad. The reward function is sparse and binary, which gives a reward of 0 if the goal has been achieved and a reward of -1 if the task failed. The agents are running at a time step of 0.04s. The policies are trained with the Message Passing Interface (MPI) [39], a parallel training tool that can run multiple DRL training threads to accelerate the experience collection process. The PC hardware for training includes an Intel 12900K, an Nvidia RTX3080ti, and 64 GB of RAM. Most hyperparameters are from [26], but with changes to the number of MPI workers to 4, total epoch to 400, cycles per epoch to 25, and batches per cycle to 25 for less training time.

### B. Evaluation Metrics

The following configurations were implemented for the ablation study and compared with baselines:
1) MAGCLA with SHER (MAGCLA+SHER)
2) MAGCLA with HER (MAGCLA+HER)
3) MADDPG with SHER (MADDPG+SHER)
4) A single DDPG agent with HER (DDPG+HER)

These experimental setups allow us to compare and test our two separate contributions, MAGCLA and SHER. Comparing 1) and 2) evaluate the improvements of the SHER compared to the HER. Likewise, we compare MAGCLA to MADDPG with 1) and 3). We also compare the improvements MAGCLA has to the conventional centralized control with 2) to 4). Lastly, we compare both our contributions in 1) to the centralized agent in 4). Significance analysis will be applied to the results.

During the training process of the above methods, the testing set contains unlimited trials with target positions that are randomly generated within the range of $(-\pi, \pi)\ rad$. Instead of logging the episode reward and average reward, the task success rate of the target policy is recorded at the end of each epoch for a direct and unbiased comparison. The success rate is the percentage of successful cases in a validation set with 50 trials. Each trial has randomly generated initial, and target poses. The success rate will be logged every 20 epochs. Each configuration is trained 3 times, each time with 400 epochs, to obtain the statistical results. A testing set is also generated, containing 100 trials with random target poses and initial poses unseen in training and validation. The testing set is reused in all evaluations for a reproducible comparison. We focused on evaluating two metrics:

1) **Method Generalizability.** We want to evaluate the generalizability of our method in different tasks. The measure is the success rate of the trained policies in the testing set. The highest success rate configuration has the best generalizability since it is the main objective of in-hand manipulation.

2) **Policy Adaptability.** The adaptability of the trained policy was evaluated with simulated robot malfunction tests. In each robot malfunction test, one agent for the multi-agent setups was manually disabled, or the corresponding finger/wrist for the one-policy setup was disabled. The malfunctioning agent was not removed; rather, it was manually set to a fully open state. The disabled wrist was set to a neutral position. Such a setting maintains the state space of each training configuration and minimizes the disturbance caused by the disabled agent. Thus, there are 6 robot malfunction tests. For simplicity, they are denoted as *no wrist, no thumb, no index, no middle, no ring, and no little*. The success rate in each malfunction test will be recorded. The percentage of performance reduction will be calculated compared with the original success rate. The configuration with the lowest average performance reduction has the best adaptability. The adaptability evaluation helps further analyze the importance of each agent for the in-hand manipulation task. The agent that causes higher performance reduction is more important.

## V. RESULTS AND DISCUSSION

### A. Training Process

The results of the training process are shown in Fig 5. Overall, the rotating block task is more difficult than the rotating egg task. The block's plane surface and sharp edges easily cause slippery movement and object falling. In contrast, the egg object has a symmetric shape and smooth curved surface, making the grasping more stable and easier to manipulate.

The multiple training processes in both tasks show consistent learning curves for all configurations. Overall, the single-agent baseline DDPG+HER (red) has the highest learning speed and converged success rate compared to all multi-agent strategies. The multi-agent baseline

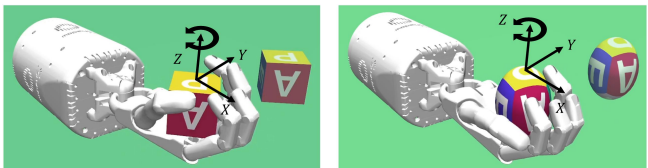

Fig. 4. Two in-hand manipulate tasks are used: block, egg. The task is to manipulate the object around the Z axis to achive an randomly generated target pose (shown on the right of the hand).

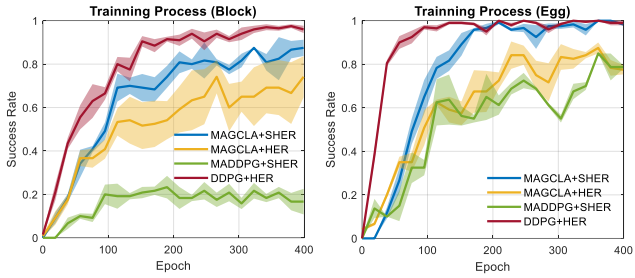

Fig. 5. The training process in block and egg rotating tasks. With SHER, MAGCLA (blue) achieved comparable learning speed and converged success rate as the single-agent DDPG (red) approach. Without SHER, MAGCLA (yellow) had lower learning speed and success rate. The conventional multi-agent approach (MADDPG in green) has the lowest learning speed and difficulty to find an optimal policy.

MADDPG+SHER (green) has the lowest learning speed and converged success rate. MAGCLA's learning speed and success rate (blue) are higher than the baseline MADDPG (green) when both with SHER. MAGCLA+SHER (blue) achieved comparable learning speed and success rate as the single agent approach (red).

The multi-agent approaches' lower learning speed is explained in [37] and [40]. The multiple agents need more exploration in the early stage to collect experience for cooperative learning. Comparing MAGCLA+SHER (blue) and MADDPG+SHER (green) proves that the extra local observation for the actor can benefit the training. Comparing MAGCLA+SHER and MAGCLA+HER (yellow) confirms that synchronizing experience replay across all agents improves learning efficiency.

*B. Generalizability, Adaptability, Importance Evaluation*

The results of the generalizability, adaptability, and agent importance are shown in Table. I. Since the performance for the MAGCLA in the no-ring test is never reduced, we will not list the entries in Table. I. The highest success rate among the trained policies for each configuration is logged.

*Method Generalizability:* Overall, MAGCLA+SHER achieved the highest success rate in both tasks showing higher generalizability than other methods. The DDPG method achieved the second lowest success rate, which means it is not generalizable in different tasks. MADDPG converged to a bad policy in the block task and the lowest success rate in the egg task, which means it has the lowest generalizability.

*Policy Adaptability:* All configurations cannot adapt to the no-wrist test since they failed in most trials. Thus, we do not consider it when calculating the average performance reduction. Overall, the proposed MAGCLA achieved the lowest average performance reduction compared to MADDPG and single DDPG. Specifically, MAGCLA+SHER achieved the lowest performance reduction in most tests except the no-little test, in which HER performs better than SHER. The reason is that the individual HER reduces the fingers' dependency on each other, providing more chances for the ring agent to learn manipulation ability, which accommodates the no-little test. While the synchronized experience replay in SHER makes the agents rely more on each other, the ring agent only learns to keep the object in hand.

*Agent Importance*: For all configurations and tasks, the wrist is the most important agent, which is reasonable as it controls the palm movement and contributes to most of the cooperative movement with the fingers. The results show two levels of importance for the fingers. The first level is the thumb and little fingers, as they have a higher importance level because their malfunctions cause higher performance reduction. The second level contains the index, middle, and ring fingers, which cause lower performance reduction. The reason is that during the manipulation, the thumb and little finger take more responsibilities to rotate the object while the remaining fingers prevent the object from falling, which are redundant to each other. The results also show that MADDPG and DDPG have much different finger importance in block and egg tasks. However, MAGCLA has consistent finger importance when comparing the results in block and egg tasks, which demonstrate MAGCLA's higher generalizability in similar tasks with different objects.

*C. Configuration Significance Analysis*

Table II shows the configurations' significance analysis. For generalizability, the analysis applies to the original success rate. For adaptability, the analysis applies to the average performance reduction of all finger agents (wrist is not considered). We chose the N-1 Chi Squared two-tailed test and considered a *p*-value less than 0.05 significant. MAGCLA shows statistical significance in generalizability and adaptability compared to the multi-agent method (MADDPG) and single-agent method (DDPG). Compared to HER, SHER shows less significance in improving generalizability and adaptability, which is reasonable because SHER mainly aims to improve training speed. The overall results show that MAGCLA contributes more to the improvement in generalizability and adaptability.

TABLE I. EVALUATION OF THE GENERALIZABILITY, ADAPTABILITY AND AGENT IMPORTANCE

| | Block Rotating Task | | | | | | | | | | | | |
|---|---|---|---|---|---|---|---|---|---|---|---|---|---|
| | Gen | Ada | | | | | | | | | | | |
| | sr | No Thumb | | No Index | | No Middle | | No Ring | | No Little | | No Wrist | | Ave rd |
| | | sr | rd | sr | rd | sr | rd | sr | rd | sr | rd | sr | rd | |
| MAGCLA+SHER | .91* | .38 | .58 | .78 | .14 | .69 | .24 | .94 | ↘ | .28 | .69 | .05 | .95 | .33 |
| MAGCLA+HER | .82 | .34 | .59 | .59 | .28 | .62 | .24 | .84 | ↘ | .47 | .43 | .06 | .93 | **.32^** |
| MADDPG+SHER | .21 | .13 | .38 | .08 | .62 | .11 | .48 | .18 | .14 | .05 | .76 | .01 | .95 | .48 |
| DDPG+HER | .76 | .20 | .74 | .44 | .42 | .54 | .29 | .74 | .03 | .11 | .86 | .01 | .99 | .47 |
| | Egg Rotating Task | | | | | | | | | | | | |
| MAGCLA+SHER | **.95** | .48 | .49 | .89 | .06 | .71 | .25 | .92 | .03 | .36 | .62 | .04 | .96 | **.29** |
| MAGCLA+HER | .87 | .43 | .51 | .66 | .24 | .54 | .38 | .80 | .08 | .42 | .52 | .02 | .98 | .34 |
| MADDPG+SHER | .76 | .31 | .59 | .24 | .68 | .27 | .64 | .37 | .51 | .25 | .67 | .03 | .96 | .62 |
| DDPG+HER | .83 | .36 | .57 | .48 | .42 | .56 | .33 | .28 | .66 | .15 | .82 | .06 | .93 | .56 |

*The highest success rate in generalizability evaluation is bolded, ^ the lowest average performance reduction percentage in adaptability test is bolded. *Gen* means generalizability, *Ada* means adaptability, *sr* means success rate, and *rd* means percentage of performance reduction.

TABLE II. SIGNIFICANCE ANALYSIS (TWO TAILED P-VALUE)

| | Block Rotating Task | | | | | | |
|---|---|---|---|---|---|---|---|
| | Gen | | | | Ada | | |
| | a | b | c | d | a | b | c | d |
| a* | ╲ | 6e-2 | **0e+0** | **4e-3** | ╲ | 8e-1 | **3e-2** | **4e-2** |
| b | 6e-2 | ╲ | **0e+0** | 3e-1 | 8e-1 | ╲ | **1e-2** | **2e-2** |
| c | **0e+0**^ | **0e+0** | ╲ | 0e+0 | **3e-2** | **1e-2** | ╲ | 9e-1 |
| d | **4e-3** | 3e-1 | 0e+0 | ╲ | **4e-2** | **2e-2** | 9e-1 | ╲ |
| | Egg Rotating Task | | | | | | |
| a | ╲ | **5e-2** | **1e-4** | **7e-3** | ╲ | 4e-1 | **3e-6** | **1e-4** |
| b | **5e-2** | ╲ | **5e-2** | 4e-1 | 4e-1 | ╲ | **8e-5** | **2e-3** |
| c | **1e-4** | **5e-2** | ╲ | 2e-1 | **3e-6** | **8e-5** | ╲ | 4e-1 |
| d | **7e-3** | 4e-1 | 2e-1 | ╲ | **1e-4** | **2e-3** | 4e-1 | ╲ |

*For simplicity, notations in Table II are: (a) MAGCLA+SHER, (b) MAGCLA+HER, (c) MADDPG+SHER, (d) DDPG+HER. ^all significant *p*-values are bolded.

## D. Finger Cooperation Analysis

The data with MAGCLA+SHER and DDPG+HER in the egg rotating task was used to analyze the finger cooperative behaviors because its round and continuous shape presents clean and clear action pattern.

The keyframes are shown in Fig. 6. More visualization can be found in the video attachment. Overall, both methods show three stages of behavior: preparation, rotation, and stabilization. In the preparation stage, the robot hand adjusts the object to a comfortable pose for rotation. In the rotation stage, the robot hand applies periodic actions to rotate the object. In the stabilization stage, the robot hand readjusts the object for a stable grasping. Specifically, MAGCLA+SHER shows that the thumb and little finger try to keep in contact with the egg and apply continuous and conjugate force to rotate the egg with a consistent speed, which is called gaiting [4] (Fig. 6a). The potential reason for the gaiting behavior is that in multi-agent approaches, the local observation of the actor leads to cooperative behaviors that rely on fewer fingers. This behavior reduces the rotation speed but improves the stability. The DDPG+HER agent tends to throw up the egg with quick wrist movement, lets it freely rotate without contact, and catches it when it drops, which we call tossing. Such a movement relies on the cooperation between the wrist to adjust the palm pose to make the egg rotate under the gravitational effect or applies instant impulses with fingers to create inertial motion for the rotation. The fingers maintain a loose grasp to avoid falling when the egg rotates and catch it once it descends. The global observation of both actor and critic helps it learn behaviors that rely on more fingers, which can rotate the object faster but more unstable. These two different behaviors explain the lower generalizability of the single-agent approach, whose aggressive policy takes fewer steps to complete the task compared to the multi-agent policies but increases the probability of the object falling in unseen trials; thus, the multi-agent approaches have better generalizability in unseen trials, with the sacrifice of time consumption.

Fig. 7 shows the plot of the object position in the X, Y, and Z axes and the rotation angle and speed on the Z axes in the same trial. The plots show periodic movement for both methods. The X and Y position plots show that MAGCLA needs more steps in the preparation stage to adjust the object position. Then it needs more periodic actions in the rotation stage but less adjustment in the stabilization stage than the single agent policy. The Z position plot shows that the single DDPG agent has more periodic movement than MAGCLA, which means the single agent relies more on the wrist movement to achieve the target position, corresponding to the tossing movement. The Z rotation angle shows that the single policy took fewer steps to achieve the target because it has a higher rotational speed, as shown in the velocity plot.

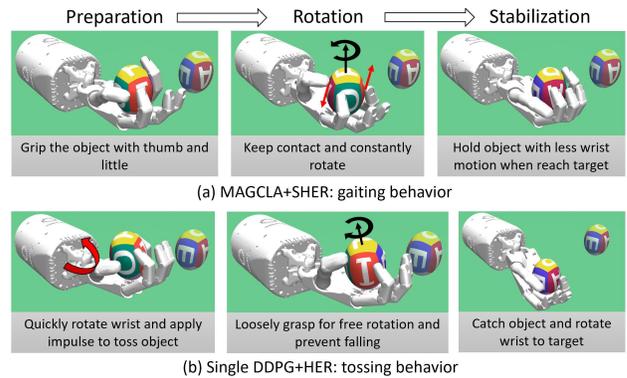

Fig. 6. The keyframes of the manipulation process in the egg rotating task. (a) MAGCLA+SHER learned conservative gaiting behavior that keeps contact on the egg and slowly rotates. (b) DDPG+HER learned aggressive tossing behavior that tosses the object to let it freely rotates.

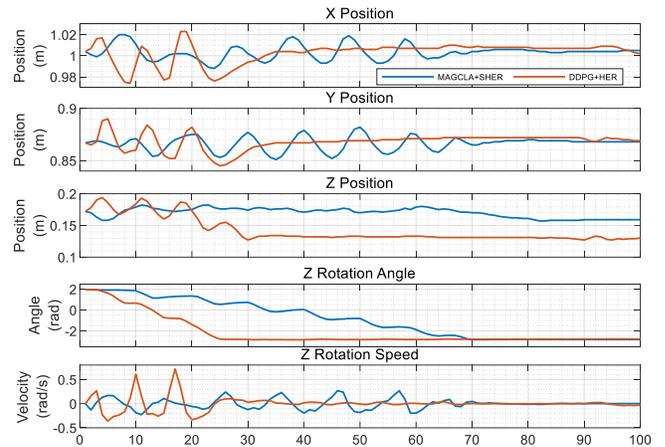

Fig. 7. The object position in X, Y, Z axes and the object rotation in Z axis. in a single trial. The plots show periodic movement for both methods. Compared to single DDPG, MAGCLA needs more steps in the preparation stage and has lower rotation speed during rotation stage, but it makes less adjustments with the wrist (Z position), making the manipulation more stable.

## VI. CONCLUSION

This work first solves the in-hand manipulation task with a multi-agent DRL setup and presents the MAGCLA approach, providing additional observation of the neighbor's action to the actor. The experiment results show that the MAGCLA approach has higher generalizability in different tasks when tresting with unseen instance. and trained policies have better adaptability to keep performance in robot malfunction. We also developed the SHER approach, which synchronizes the experience across all agents to improve the learning efficiency of MAGCLA to reach a comparable training speed to the single-agent approach. In this work, we evaluate the MAGCLA-SHER method in two in-hand manipulation tasks, but it is the highest potential to be more applicable in more in-hand manipulation tasks and other multi-agent tasks that require more cooperative behaviors. Our future work will focus on embedding our methodin more DRL algorithms and studying their feasibility in real-world applications.